%% file: iclr2020_conference.tex
\documentclass{article} % For LaTeX2e
\usepackage{iclr2020_conference,times}

\input{math_commands.tex}

\usepackage{hyperref}
\usepackage{url}
\usepackage{graphicx}
\newcommand{\comment}[1]{}

\title{Unsupervised Pidgin Text Generation By Pivoting English Data and Self-Training}

\author{Ernie Chang$^1$, David Ifeoluwa Adelani$^2$, Xiaoyu Shen$^2$ \& Vera Demberg$^1$
\\
$^1$Department of Language Science and Technology, Saarland University, Germany\\
$^2$Spoken Language Systems (LSV), Saarland Informatics Campus, Germany\\
\\
\texttt{\{cychang,vera\}@coli.uni-saarland.de} \\
}

\iclrfinalcopy % Uncomment for camera-ready version, but NOT for submission.
\begin{document}

\maketitle

\begin{abstract}
West African Pidgin English is a language that is significantly spoken in West Africa, consisting of at least 75 million speakers.
Nevertheless, proper machine translation systems and relevant NLP datasets for pidgin English are virtually absent.
In this work, we develop techniques targeted at bridging the gap between Pidgin English and English in the context of natural language generation.
%As a proof of concept, we explore the proposed techniques in the area of data-to-text generation. 
By building upon the previously released monolingual Pidgin English text and parallel English data-to-text corpus, we hope to build a system that can automatically generate Pidgin English descriptions from structured data.
We first train a data-to-English text generation system, before employing techniques in unsupervised neural machine translation and self-training to establish the Pidgin-to-English cross-lingual alignment.
The human evaluation performed on the generated Pidgin texts shows that, though still far from being practically usable, the pivoting + self-training technique improves both Pidgin text \emph{fluency} and \emph{relevance}.
\end{abstract}

\section{Introduction}

Pidgin English is one of the the most widely spoken languages in West Africa with roughly 75 million speakers estimated in Nigeria; and over 5 million speakers estimated in Ghana \citep{ogueji2019pidginunmt}.
\footnote{Though variants of Pidgin English are abound, the language is fairly uniform across the continent. 
In this work, we directed our research to the most commonly spoken variant of West African Pidgin English $-$ the Nigerian Pidgin English.}
While there have been recent efforts in popularizing the monolingual Pidgin English as seen in the BBC Pidgin\footnote{ \url{https://www.bbc.com/pidgin} },
it remains under-resourced in terms of the available parallel corpus for machine translation.
Similarly, this low-resource scenario extends to other domains in natural language generation (NLG) such as summarization, data-to-text and so on~\citep{lebret2016neural,hong2019improving,de2018generating,chang2021neural,chang2021jointly}
$-$ where Pidgin English generation is largely under-explored.
The scarcity is further aggravated when the pipeline language generation system includes other sub-modules that computes semantic textual similarity~\citep{zhuang2017neobility}, which exists solely in English.

Previous works on unsupervised neural machine translation for Pidgin English constructed a monolingual corpus \citep{ogueji2019pidginunmt}, 
and achieved a BLEU score of 5.18 from English to Pidgin.
However, there is an issue of domain mismatch between down-stream NLG tasks and the trained machine translation system.
This creates a caveat where the resulting English-to-Pidgin MT systems (trained on the domain of news and the Bible) cannot be directly used to translate out-domain English texts to Pidgin. An example of the English/pidgin text in the restaurant domain~\citep{novikova2017e2e} is displayed in Table ~\ref{fig:sample}.

\begin{table*}[h]
\begin{tabular}{l|l}
English  & There is a pub Blue Spice located in the centre of the city that provides Chinese food.                 \\ \hline
Pidgin  & Wan pub blue spice dey for centre of city wey dey give Chinese food.
                  \\ 
\end{tabular}
    \caption{ Sample parallel English-Pidgin text from the restaurant domain.
    }
    \label{fig:sample}
\end{table*}
\comment{
\begin{table*}[h]
\begin{tabular}{l|l}

English text & What will the weather be like in New York City at 5pm?                  \\ \hline
Pidgin text  & What will the weather be like in New York City at 5pm?                  \\ 
\end{tabular}
    \caption{ Sample utterances from the Weather dataset \cite{constrained}.
    }
    \label{fig:sample2}
\end{table*}}

Nevertheless, we argue that this domain-mismatch problem can be alleviated by using English text in the target-domain as a pivot language \citep{guo2019zero}.
To this end, we explore this idea on the task of neural data-to-text generation which has been the subject of much recent research. 
Neural data-to-Pidgin generation is essential in the African continent especially given the fact that many existing data-to-text systems are English-based e.g. Weather reporting systems \citep{sripada2002sumtime,belz2008automatic}.
This work aims at bridging the gap between many of these English-based systems and Pidgin by training an \emph{in-domain} English-to-pidgin MT system in an unsupervised way.
By this means, English-based NLG systems can be locally adapted by translating the output English text into pidgin English. 
We employ the publicly available parallel data-to-text corpus E2E \citep{novikova2017e2e} \comment{and Weather \citep{constrained}} consisting of tabulated data and English descriptions in the restaurant domain.
\comment{The input data for E2E is in the form of meaning representation (MR) consisting of attributes and values paired with the target English text.}
The training of the in-domain MT system is done with a two-step process:
(1) We use the target-side English texts as the pivot, and train an unsupervised NMT ($model_{unsup}$) directly between in-domain English text and the available monolingual Pidgin corpus. 
(2) Next, we employ self-training \citep{he2019revisiting} to create augmented parallel pairs to continue updating the system ($model_{self}$).

\comment{
\section{Related Works}
\label{related}

Previous works on unsupervised neural machine translation for Pidgin English constructed a monolingual corpus \citep{ogueji2019pidginunmt}, 
and achieved a BLEU score of 7.93 from Pidgin to English and 5.18 from English to Pidgin.
However, there is largely an issue where the desired monolingual Pidgin text suffers from domain mismatch e.g. between restaurant domain in E2E and largely news domain for Pidgin monolingual data.
This creates a caveat where a trained Pidgin-English NMT is not applicable.
}

\section{Approach}
\label{approach}

First phase of the approach requires training of an unsupervised NMT system similar to  \citet{ogueji2019pidginunmt} (PidginUNMT).
Similar to \citet{ogueji2019pidginunmt}, we train the cross-lingual model using FastText \cite{bojanowski2017enriching} on the combined Pidgin-English corpus.
Next, we train an unsupervised NMT similar to \citet{lample2017unsupervised,artetxe2017unsupervised,ogueji2019pidginunmt} between them to obtain $model_{unsup}$.
Then we further utilize $model_{unsup}$ to construct pseudo parallel corpus by predicting target Pidgin text given the English input.
We augment this dataset to the existing monolingual corpus.
The self-training step involves further updating  $model_{unsup}$ on the pseudo parallel corpus and non-parallel monolingual corpus to yield  $model_{self}$.

\section{Experiments and Results}
\label{exp}

We conduct experiments on the E2E corpus~\citep{novikova2017e2e} which amounts to roughly 42k samples in the training set.
The monolingual Pidgin corpus contains 56,695 sentences and 32,925 unique words.
The human evaluation was performed on the test set (630 data instances for E2E) by averaging over scores by 2 native Pidgin speakers on both Relevance (0 or 1 to indicate relevant or not) and Fluency (0, 1, or 2 to indicate readability).
Table~\ref{fig:score} shows that $model_{self}$ outperforms direct translation (PidginUNMT) and unsupervisedly-trained model $model_{unsup}$ on \emph{relevance} and performing on par with PidginUNMT on \emph{fluency}.
We also display relevant sample outputs in Table~\ref{fig:out} at all levels of fluency.
\comment{$0$ indicates that it is gibberish, $1$ shows that it is somewhat readable, and $2$ would mean that it is perfectly fine as a Pidgin English sentence.}

\begin{table}[]
\centering
\begin{tabular}{c|c|c}
           %& \multicolumn{2}{c}{ \textbf{E2E}}    & \multicolumn{2}{c}{\textbf{Weather} }                                         \\
\textbf{System} & \textbf{Relevance} & \textbf{Fluency}  \\ \hline
PidginUNMT  & 0.038     & \textbf{0.827}  \\ 
$model_{unsup}$ & 0.319     & 0.788   \\ 
$model_{self}$ & \textbf{0.434}     & 0.814   \\ 
\end{tabular}
\caption{PidginUNMT is trained on unparallel, out-domain English and pidgin text.
$model_{unsup}$ refers to unsupervised NMT trained on in-domain English text and out-domain pidign text. $model_{self}$ further augments $model_{unsup}$ with pseudo parallel pairs obtained from self-training. 
}
    \label{fig:score}
\end{table}

\begin{table}[h]
\resizebox{0.8\textwidth}{!}{
\begin{tabular}{l|c}
\textbf{ Pidgin text}                                                                      & \textbf{Fluency} \\ \hline
Every money of money on food and at least 1 of 1 points. & 0 \\ \hline
and na one na di best food for di world. & 1       \\ \hline
People dey feel the good food but all of us no dey available. & 2       \\ 
\end{tabular}
}
\caption{
Sampled relevant (score of 1) Pidgin outputs from $model_{self}$ with various Fluency scores.
}
    \label{fig:out}
\end{table}

\section{conclusion}
\label{conclusion}

In this paper, we have shown that it is possible to improve upon low-resource Pidgin text generation in a demonstrated low-resource scenario. 
By using non-parallel in-domain English and out-domain Pidgin text along with self-training algorithm, we show that both \emph{fluency} and \emph{relevance} can be further improved. 
This work serves as the starting point for future works on Pidgin NLG in the absence of annotated data.
For future works, we will also further utilize phrase-based statistical machine translation to further improve upon current work.

\section{Acknowledgements}
\label{acknowledgements}

This research was funded in part by the SFB 248 “Foundations of Perspicuous Software Systems”. 
We sincerely thank the anonymous reviewers for their insightful comments that helped us to improve this paper.

\comment{
\subsubsection*{Author Contributions}
If you'd like to, you may include  a section for author contributions as is done
in many journals. This is optional and at the discretion of the authors.

\subsubsection*{Acknowledgments}
Use unnumbered third level headings for the acknowledgments. All
acknowledgments, including those to funding agencies, go at the end of the paper.
}

\bibliography{iclr2020_conference}
\bibliographystyle{iclr2020_conference}

\comment{
\appendix
\section{Appendix}
You may include other additional sections here. 
}

\end{document}

%% file: math_commands.tex
\usepackage{amsmath,amsfonts,bm}

\def\eqref#1{equation~\ref{#1}}
\def\1{\bm{1}}

\DeclareMathAlphabet{\mathsfit}{\encodingdefault}{\sfdefault}{m}{sl}
\SetMathAlphabet{\mathsfit}{bold}{\encodingdefault}{\sfdefault}{bx}{n}